\documentclass[letterpaper]{article} 
\usepackage{aaai2026}  
\usepackage{times}  
\usepackage{helvet}  
\usepackage{courier}  
\usepackage[hyphens]{url}  
\usepackage{graphicx} 
\usepackage{natbib}  
\usepackage{caption} 
\usepackage{amsmath}      
\usepackage{amssymb}      
\usepackage{adjustbox} 
\usepackage{enumitem} 
\usepackage[table]{xcolor}
\usepackage{adjustbox}
\usepackage{booktabs} 
\usepackage{subcaption}
\usepackage{algorithm}
\usepackage{algpseudocode}
\usepackage{float}
\usepackage{booktabs}
\usepackage{adjustbox}
\usepackage{newfloat}
\usepackage{listings}
\DeclareCaptionStyle{ruled}{labelfont=normalfont,labelsep=colon,strut=off} 
\floatstyle{ruled}
\newfloat{listing}{tb}{lst}{}
\floatname{listing}{Listing}
\title{BLM-Guard: Explainable Multimodal Ad Moderation with Chain-of-Thought and Policy-Aligned Rewards}
\author {
    Yiran Yang\textsuperscript{\rm 1,2}\equalcontrib\thanks{Work done during an internship at Kuaishou Technology.},
    Zhaowei Liu\textsuperscript{\rm 1}\equalcontrib,
    Yuan Yuan\textsuperscript{\rm 1},
    Yukun Song\textsuperscript{\rm 1,2},
    Xiong Ma\textsuperscript{\rm 1},
    Yinghao Song\textsuperscript{\rm 1},
    Xiangji Zeng\textsuperscript{\rm 1},
    Lu Sun\textsuperscript{\rm 1},
    Yulu Wang\textsuperscript{\rm 1},
    Hai Zhou\textsuperscript{\rm 1},
    Shuai Cui\textsuperscript{\rm 1,3},
    Zhaohan Gong\textsuperscript{\rm 1},
    Jiefei Zhang\textsuperscript{\rm 1}\thanks{Project Leader}
}

\affiliations {
    \textsuperscript{\rm 1}Kuaishou Technology, 
    \textsuperscript{\rm 2}Beijing University of Posts and Telecommunications, 
    \textsuperscript{\rm 3}Shandong University\\
    yyr2023110885@bupt.edu.cn, songyukun@bupt.edu.cn, 202315156@mail.sdu.edu.cn, \\
    \{liuzhaowei03, yuanyuan06, maxiong, songyinghao03, zengxiangji, sunlu05, wangyulu05, zhouhai, gongzhaohan, zhangjiefei\}@kuaishou.com
}

\usepackage{bibentry}

\begin{document}

\maketitle

\begin{abstract}
Short-video platforms now host vast multimodal ads whose deceptive visuals, speech and subtitles demand finer-grained, policy-driven moderation than community safety filters. We present BLM-Guard, a content-audit framework for commercial ads that fuses Chain-of-Thought reasoning with rule-based policy principles and a critic-guided reward. A rule-driven ICoT data-synthesis pipeline jump-starts training by generating structured scene descriptions, reasoning chains and labels, cutting annotation costs. Reinforcement learning then refines the model using a composite reward balancing causal coherence with policy adherence. A multitask architecture models intra-modal manipulations (e.g., exaggerated imagery) and cross-modal mismatches (e.g., subtitle–speech drift), boosting robustness. Experiments on real short-video ads show BLM-Guard surpasses strong baselines in accuracy, consistency and generalization.
\end{abstract}

\section{Introduction}

With the rise of generative AI and the rapid growth of short-video platforms like \textit{TikTok} and \textit{Instagram Reels}, creating multimodal advertising content has become easier and cheaper than ever~\cite{qi2022fakesvmultimodalbenchmarkrich,yeh2024t2vsmeetvlmsscalable}. Brands and individuals can now swiftly produce videos that combine visuals, speech, and text with minimal cost. While this boosts monetization efficiency and creative diversity, it also poses new challenges for content moderation.

Unlike general community moderation—typically focused on coarse risks like violence or nudity~\cite{helff2025llavaguardopenvlmbasedframework,inan2023llamaguardllmbasedinputoutput}—ad moderation demands \textit{fine-grained, policy-driven} compliance checks. Violations are often subtle or disguised, including exaggerated claims, misleading cues, or rule evasion~\cite{goldstein2023generativelanguagemodelsautomated}, and frequently manifest as \textit{modality misalignments}—e.g., visually truthful but verbally deceptive content, benign captions with provocative imagery, or mismatched subtitles and audio~\cite{guo2024moderatingillicitonlineimage}.

However, existing moderation approaches—such as static rule-based filters or general-purpose vision–language models (VLMs)—struggle in high-risk, multimodal ad settings due to three key limitations:
(i) limited cross-modal causal reasoning~\cite{ahn-etal-2024-tuning};
(ii) poor adaptability to policy drift~\cite{lu2025vlmpolicycommonlawcontent}; and
(iii) lack of task-specific reasoning for nuanced commercial risks~\cite{liu2025guardreasonervlsafeguardingvlmsreinforced}.
Most prior work targets generic harm detection~\cite{qu2024unsafebenchbenchmarkingimagesafety}, offering limited support for regulation-sensitive ad content.

We propose \textbf{BLM-Guard}, a moderation framework for policy-sensitive short-video ads, featuring three key innovations:
(1) An \textit{Interleaved-modal Chain-of-Thought (ICoT)} pipeline \cite{gao2025interleavedmodalchainofthought} that integrates visual grounding with causal reasoning \cite{wang2023selfinstructaligninglanguagemodels,xia2025msralignpolicygroundedmultimodalalignment} for explainable decisions.
(2) A self-adaptive GRPO reinforcement learning framework with hybrid rewards fusing rule correctness and \textit{principle-based self-consistency} \cite{chen2025grpocareconsistencyawarereinforcementlearning} to handle policy shifts.
(3) A multi-task architecture modeling intra-modal and cross-modal violations to enhance robustness and generalization.

\begin{figure*}
    \centering
    \includegraphics[width=0.95\linewidth]{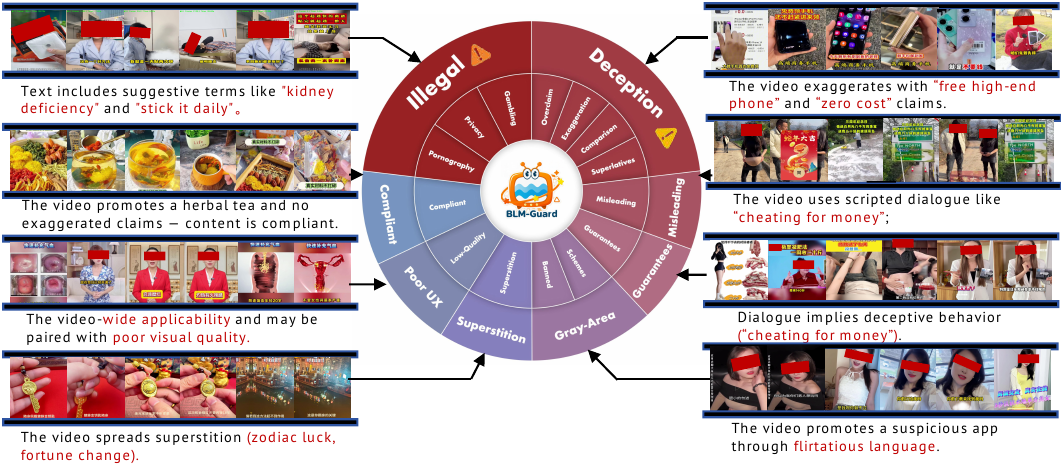}
    \caption{%
        \textbf{BLM-Guard Benchmark Taxonomy.} Our benchmark organizes commercial short-video ads into a hierarchical risk taxonomy with seven core violation scenarios and fine-grained subtypes. Each node reflects a policy-sensitive violation type (e.g., income exaggeration, privacy leak, feudal superstition), and is further associated with a severity level (high, medium, low). This structure supports interpretable supervision and enables fine-grained evaluation of model performance on diverse and nuanced moderation cases. A dedicated “No Risk” category is also included to balance risk distribution.%
        }

    \label{fig:benchmark}
\end{figure*}

To support systematic evaluation, we release the \textbf{BLM-Guard Benchmark}, a real-world dataset of short-video ads annotated with a three-level risk taxonomy aligned with platform policies:
(i) \textit{risk scenario} (e.g., illegal content, misleading marketing);
(ii) \textit{violation type} (e.g., income exaggeration, privacy exposure);
(iii) \textit{risk severity} (high, medium, low).
This hierarchical design supports interpretable supervision and rigorous policy-aligned evaluation. A visual summary is shown in Figure~\ref{fig:benchmark}.

\paragraph{Our main contributions are:}
\begin{itemize}
\item We introduce \textbf{BLM-Guard Benchmark}, a real-world dataset for ad moderation, structured across seven risk scenarios and fine-grained violation types, enabling policy-grounded evaluation.
\item We propose \textbf{BLM-Guard}, a multimodal moderation framework combining rule-driven ICoT reasoning, consistency-aware reinforcement learning, and multi-task modeling to ensure policy-compliant and explainable decisions.
\end{itemize}

\section{Task Formulation and BLM-Guard Benchmark}

\subsection{Task Definition}
\label{sec:task-def}

We formulate the task as policy-aligned moderation of commercial short-video ads. Given a video $\mathbf{v} = \langle f_1, f_2, \dots, f_T \rangle$ composed of multimodal frames (visuals and ASR transcripts), and a policy set $\mathcal{P} = \{\pi_1, \dots, \pi_n\}$ from platform guidelines, the goal is to determine for each policy $\pi_i$ whether it is violated, and to generate an interpretable reasoning trace.

Formally, the moderation model $G_{\text{reasoner}}$ outputs a decision $\hat{Y}_i \in \{0,1\}$ and a reasoning chain $R_i$:
\begin{equation}
\bigl\{(\hat{Y}_i, R_i)\bigr\}_{i=1}^{n} = G_{\text{reasoner}}(\mathbf{v}, \mathcal{P}),
\end{equation}
where $\hat{Y}_i = 1$ denotes a violation. Each $R_i$ follows a structured path:
\textit{Scenario Identification} $\rightarrow$ \textit{Violation Typing} $\rightarrow$ \textit{Rule Matching} $\rightarrow$ \textit{Decision}.

This structure supports explainable moderation and facilitates reward modeling for reinforcement learning.

\subsection{BLM-Guard Benchmark}
\label{sec:benchmark}

To support structured moderation training, we introduce the \textbf{BLM-Guard Benchmark}, a short-video dataset with three-level policy labels (Figure~\ref{fig:benchmark}).
\begin{itemize}[leftmargin=1em]
    \item \textbf{Level 1: Severity} — High, Medium, or Low, reflecting potential legal, reputational, or user experience risks.
    \item \textbf{Level 2: Scenario} — e.g., illegal content, false marketing, misleading operations, etc.
    \item \textbf{Level 3: Violation Type} — e.g., income exaggeration, privacy leak, feudal superstition, and others.
\end{itemize}

This hierarchical structure enables fine-grained, policy-aligned supervision and supports nuanced detection of multimodal violations.

\subsection{Data Construction}
\label{sec:data}

We collect 5–45s commercial ads from a short-video platform, covering domains like e-commerce, health, education, and lifestyle. The dataset includes:
\begin{enumerate}[label=\textbf{\arabic*.}, leftmargin=1.5em]
    \item \textbf{Expert-Audited Ads:} Videos labeled by professional moderators with verified risk types.
    \item \textbf{Rule-Triggered Ads:} Samples flagged by rule-based moderation systems.
    \item \textbf{High-Impression Compliant Ads:} Popular videos with low complaint rates to improve generalization.
\end{enumerate}

Each video is annotated with its risk severity, scenario, and violation type. A subset contains structured reasoning traces to support both supervised learning and interpretable reward-based training.

\begin{figure*}[t]
    \centering
    \includegraphics[width=1\linewidth]{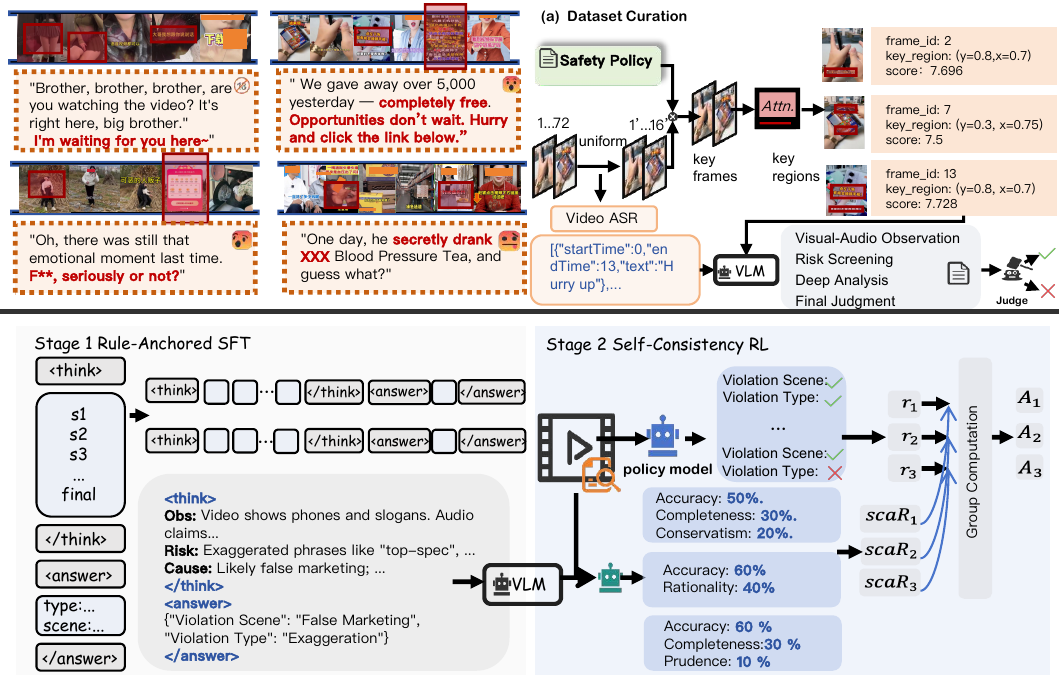}
    \caption{Our method adopts a progressive two-stage pipeline for policy-controllable content moderation. In Stage 1 (Rule-driven SFT Cold Start), we synthesize structured visual-language Chain-of-Thought (ICoT) data via keyframe selection and multi-step prompting using InternVL. This enables supervised fine-tuning with rule-anchored causal supervision. In Stage 2 (Self-adaptive GRPO Reinforcement Learning), we apply a safety-aware data curation strategy and propose a Self-Adaptive Critique Reward (SACR) to dynamically evaluate reasoning outputs. The model is optimized using a modified Group-wise Relative Policy Optimization (GRPO) algorithm with token-level normalization and dynamic sampling. This multi-stage process enables the model to first learn fine-grained compliance reasoning and then refine its moderation behavior through adaptive, reward-driven training.}
    \label{fig:method}
\end{figure*}

\section{Methodology}
\label{sec:method}

As shown in Fig.~\ref{fig:method}, we address the compliance and controllability challenges of short-video ad moderation via a two-stage pipeline. To overcome the limitations of single-stage tuning in policy coverage and interpretability, we decompose the training into:

\begin{itemize}[leftmargin=1.5em, itemsep=0pt]
    \item \textbf{Rule-guided SFT:} Injecting policy priors and reasoning skills into the VLM through Interleaved-modal Chain-of-Thought (ICoT) supervision.
    \item \textbf{Self-adaptive GRPO:} Optimizing behavior via Group-wise Relative Policy Optimization guided by dynamic rule-aware rewards.
\end{itemize}

This design enables the model to first acquire structured compliance reasoning and then adaptively refine moderation accuracy against evolving risk patterns.

\subsection{Cold Start Strategy: Rule-Guided Causal Supervision}
\label{sec:cold-start}

To provide an effective initialization for subsequent reinforcement learning, we propose a two-stage strategy that constructs structured visual reasoning data based on safety rules:

\begin{itemize}
    \item \textbf{Stage-1:} Keyframe and Region Selection via Risk-Prompt Anchored Coarse Filtering
    \item \textbf{Stage-2:} Interleaved Multi-stage CoT Generation using Vision-Language Models
\end{itemize}

\paragraph{Keyframe and Region Extraction.}
We employ a two-step selection strategy to extract informative frames and salient regions that represent high-risk cues:

\begin{enumerate}
    \item \textbf{CLIP-based Prompt Similarity.} For a video with $N=16$ uniformly sampled frames $\{f_i\}_{i=1}^{16}$, we compute their semantic similarity to a set of $K=7$ predefined risk prompts (e.g., ``false marketing'', ``illegal content'') using CLIP-ViT-L/14 \cite{radford2021learningtransferablevisualmodels}. Denote the visual embedding of frame $i$ as $\mathbf{v}_i \in \mathbb{R}^d$ and prompt embedding $k$ as $\mathbf{t}_k \in \mathbb{R}^d$. The maximum similarity score of frame $i$ is:

    \begin{equation}
        s_i = \max_{k} \left( \frac{\mathbf{v}_i^\top \mathbf{t}_k}{\|\mathbf{v}_i\| \cdot \|\mathbf{t}_k\|} \right)
    \end{equation}

    This results in a similarity vector $\mathbf{s} = [s_1, s_2, \dots, s_{16}] \in \mathbb{R}^{16}$.

    \item \textbf{Adaptive Keyframe Selection (AKS).} To ensure both temporal coverage and semantic diversity, we apply a hybrid \textit{BIN+TOP} strategy~\cite{tang2025adaptivekeyframesamplinglong} to select $m=3$ keyframes. Specifically, the video is evenly divided into $m$ temporal bins, and the most relevant frame is selected within each bin (BIN). If the number of selected frames is less than $m$, we supplement it with top-scoring frames across the entire video (TOP), ensuring complementary selection of globally salient frames.

    \begin{equation}
        \mathcal{K}_{\text{bin}} = \bigcup_{j=1}^{m} \left\{ \arg\max_{i \in \mathcal{S}_j} s_i \right\},\quad 
        \mathcal{S}_j = \left[ \frac{(j-1) \cdot N}{m}, \frac{j \cdot N}{m} \right)
    \end{equation}
    If $|\mathcal{K}_{\text{bin}}| < m$, we supplement it with top-ranked global scores not in $\mathcal{K}_{\text{bin}}$:
    \begin{equation}
        \mathcal{K}_{\text{final}} = \mathcal{K}_{\text{bin}} \cup \left\{ \text{Top-}(m - |\mathcal{K}_{\text{bin}}|) \right\}
    \end{equation}

    \item \textbf{Patch-level Region Extraction.} For each selected frame $f_i$, we feed it into InternViT-6B to extract patch-wise features $\mathbf{H}_i \in \mathbb{R}^{P \times d}$. The saliency score for patch $p$ is computed via L2 norm:
    \begin{equation}
        \mathrm{score}_{i,p} = \left\| \mathbf{H}_i^{(p)} \right\|_2
    \end{equation}
    The highest scoring patch is selected as the visual key region.
\end{enumerate}

Combining the selected frames and regions, we define the structured visual clue as:
\[
\mathcal{K} = \left\{ (f_i, \mathrm{region}_i) \right\}_{i=1}^3
\]

\paragraph{Interleaved Multi-stage CoT Generation.}
Given the keyframe set $\mathcal{K}$, ASR transcripts, and a predefined safety rule schema, we perform a three-stage prompting process with a frozen InternVL-3-78B model to generate structured reasoning:

\begin{itemize}
    \item \textbf{Step-1 (Observation)}: Describe the visual content, summarize ASR transcripts, and assess the consistency between modalities.
    \item \textbf{Step-2+3 (Risk Screening + Causal Analysis)}: Identify potential policy violations and analyze their underlying causes.
    \item \textbf{Step-4 (Final Verdict)}: Integrate multi-stage reasoning to reach a final compliance decision in accordance with the rule schema.
\end{itemize}

For each sample $j$, the model produces a visual-text-label triplet:
\[
\left( K^{(j)},\ T^{(j)},\ y^{(j)} \right),\quad j = 1, 2, \dots, |\mathcal{D}|
\]
where $K$ denotes the selected keyframes, $T$ the reasoning chain, and $y$ the predicted compliance label.

Internally, the model first generates intermediate reasoning tuples $(\text{obs}_i, \text{risk}_i, \text{cause}_i, \text{verdict}_i)_{i=1}^{16}$ for each frame, which are then aggregated into the final triplet $\bigl(K,\,T,\,y\bigr)$.

\subsubsection{Rule-Anchored Supervised Fine-tuning (SFT)}
\label{sec:rule-sft}

On top of the synthesized dataset, we perform supervised fine-tuning of the base VLM. The training objective consists of two components:
\[
\mathcal{L}
  = \mathcal{L}_{\mathrm{CE}}(\langle \text{answer} \rangle)
  + \lambda\,\mathrm{KL}\bigl(p_{\text{think}} \parallel p_{\text{rule}}\bigr),
\]

where the primary loss \(\mathcal{L}_{\mathrm{CE}}\) computes the cross-entropy on the \texttt{<answer>} token, encouraging accurate prediction of the compliance label. The auxiliary KL term aligns the model's reasoning distribution (\texttt{<think>}) with a rule-guided prior \(p_{\text{rule}}\), which we construct by tokenizing the target violation scene and type (e.g., ``false marketing'', ``adult content'') into a keyword set. These keywords are normalized into a soft token-level target distribution, serving as an interpretable and structured prior to guide the model's causal reasoning.

In summary, we construct a compact yet effective multimodal CoT corpus and perform rule-anchored fine-tuning to establish a solid cold-start foundation for downstream reinforcement learning.

\subsection{Reinforcement Learning: Principle-Guided Self-Consistent Optimization}
\label{sec:rl}

Building on the cold-start model trained with multimodal CoT-style supervision, we apply online reinforcement learning (RL) to further enhance the model’s reasoning ability and policy alignment in content moderation scenarios. The overall training procedure is illustrated in \textbf{Algorithm~\ref{alg:blm-guard-grpo}}, which consists of three main components:

\begin{itemize}
    \item \textbf{Data Curation}: Constructing difficult and diverse samples to improve generalization.
    \item \textbf{Reward Design}: Combining rule-based correctness, principle-guided critique, and format consistency.
    \item \textbf{Policy Optimization}: Applying GRPO with token-level normalization and dynamic sampling.
\end{itemize}

To improve generalization and sample efficiency in RL, we construct a refined dataset $\mathcal{D}_{\text{RL}}$ from the supervised set $\mathcal{D}$ through the following strategies:

\paragraph{(1) Rejection Sampling.}
We sample from $\mathcal{D}$ with high temperature for 4 passes, collecting all examples where the model fails consistently across generations as hard samples.

\paragraph{(2) Safety-Aware Concatenation.}
To increase multimodal diversity and risk density, we concatenate semantically related prompts or risky ASR segments to simulate more complex decision contexts.

\textit{Note: We use only ASR since OCR in short videos is often noisy (e.g., overlays), which harms reasoning. ASR provides more stable guidance.}

\begin{algorithm}[H]
\small
\caption{BLM-Guard: Self-Adaptive GRPO Training}
\label{alg:blm-guard-grpo}
\begin{algorithmic}
\Procedure{Train}{$\pi_\theta$, $\mathcal{R}$, $\mathcal{D}_{\text{RL}}$, $S$}
    \Comment{\textbf{Inputs}: 
    $\pi_\theta$ = moderation model (from cold-start), 
    $\mathcal{R}$ = reward model (rule + format + SCA-R), 
    $\mathcal{D}_{\text{RL}}$ = reasoning dataset, 
    $S$ = training steps, 
    $G$ = group size per input, 
    $\varepsilon$ = clip scale}

    \For{$s = 1$ to $S$}
        \For{each $x$ in batch from $\mathcal{D}_{\text{RL}}$}

            \State \textbf{Phase 1: Response Generation and Reward}
            \State Generate $G$ responses $\{o_i\}_{i=1}^G \sim \pi_\theta(x)$
            \For{$i = 1$ to $G$}
                \State $r^{\text{rule}}_i \gets$ rule match score
                \State $r^{\text{format}}_i \gets$ format compliance score
                \State $r^{\text{scaR}}_i \gets$ critique score from guide model
                \State $r_i \gets r^{\text{rule}}_i + r^{\text{format}}_i + r^{\text{scaR}}_i$
            \EndFor

            \If{$\operatorname{Var}(\{r_i\}) = 0$}
                \State \textbf{continue} \Comment{Skip zero-variance batch}
            \EndIf

            \State $\mu, \sigma \gets$ mean and std of $\{r_i\}$

            \State \textbf{Phase 2: GRPO Optimization}
            \For{$i = 1$ to $G$}
                \State $A_i \gets \dfrac{r_i - \mu}{\sigma}$ \Comment{Z-score advantage}
                \State $K_i \gets \dfrac{P_\theta(o_i)}{P_{\theta_{\text{old}}}(o_i)}$
            \EndFor

            \State $B_s \gets \varepsilon \cdot \dfrac{S - s}{S}$ \Comment{Annealed clip factor}

            \State Compute loss:
            \[
            \mathcal{L}_{\text{RL}} = - \frac{1}{G} \sum_{i=1}^{G}
            \min \left( K_i A_i,\; \text{clip}(K_i, 1 - B_s, 1 + B_s) \cdot A_i \right)
            \]

            \State $\theta \gets \theta - \eta \cdot \nabla_\theta \mathcal{L}_{\text{RL}}$
            \State $\theta_{\text{old}} \gets \theta$ \Comment{Update reference policy}

        \EndFor
    \EndFor
    \State \Return Optimized policy $\pi_\theta$
\EndProcedure
\end{algorithmic}
\end{algorithm}

\subsubsection{Reward Design}

We propose a hybrid reward mechanism that combines factual correctness, interpretable reasoning, and structured output format. The final reward used in RL is:

\[
r = r_{\text{rule}} +  r_{\text{format}}
 + r_{\text{scaR}}
\]
\paragraph{(1) Rule-Based Reward.}
We assign a discrete correctness reward based on whether the predicted violation \texttt{scene} and \texttt{type} match the ground truth. Each correct field contributes 0.5:
\[
r_{\text{rule}} = 
\begin{cases}
    1.0, & \text{if } \hat{y}_{\texttt{scene}} = y_{\texttt{scene}} \text{ and } \hat{y}_{\texttt{type}} = y_{\texttt{type}} \\
    0.5, & \text{if } \hat{y}_{\texttt{scene}} = y_{\texttt{scene}} \text{ and } \hat{y}_{\texttt{type}} \ne y_{\texttt{type}} \\
    0.0, & \text{otherwise}
\end{cases}
\]

\paragraph{(2) Format-Aware Reward.}
To ensure outputs follow the required structure, we assign a binary reward based on the presence of both \texttt{<think>} and \texttt{<answer>} tags:
\[
r_{\text{format}} = \mathbb{I}[\texttt{<think>} \in \hat{y} \,\land\, \texttt{<answer>} \in \hat{y}]
\]

\paragraph{(3) SCA-R: Self-Consistency and Adaptive Reward.}
To address reward misalignment caused by policy drift, we propose \textbf{SCA-R}, a self-consistency reward guided by scenario-adaptive critique.

For each (input, response) pair, a guide model acts as a rating expert. It receives the \texttt{<think>} trace, ground-truth labels, and moderation guidelines, then dynamically constructs scoring principles $\mathcal{P} = \{p_k\}$ (e.g., causal clarity, risk attribution), each with weight $w_k$.

The model conducts step-by-step critique and assigns a score in \{0, 0.5, 1\}, where 1 indicates both reasoning and conclusion are aligned with policy.

\[
r_{\text{scaR}} = \sum_{k=1}^{K} w_k \cdot \mathrm{score}_{p_k}(\hat{y})
\]

This adaptive reward enhances interpretability while maintaining alignment under evolving moderation standards.

\subsubsection{Policy Optimization with GRPO}

We adopt Group Relative Policy Optimization (GRPO) and introduce several modifications for improved stability and sample efficiency:

\paragraph{Token-Level Normalization.}
Unlike standard GRPO that normalizes loss at the sequence level, we compute the objective at the token level to mitigate reward bias due to varying output lengths.

\paragraph{Group Advantage.}
For each input, we sample $G$ responses and compute relative advantage:

\begin{equation}
A_i = \frac{r_i - \text{mean}(\{r_j\}_{j=1}^{G})}{\text{std}(\{r_j\}_{j=1}^{G})}
\end{equation}

\paragraph{Clipped Policy Loss.}
The final RL objective is:

\begin{equation}
\mathcal{L}_{\text{RL}} = -\mathbb{E} \left[ \frac{1}{G} \sum_{i=1}^{G} \min \left( K_i, \text{clip}(K_i, 1 - B, 1 + B) \right) \cdot A_i \right]
\end{equation}

where the policy ratio is:

\begin{equation}
K_i = \frac{P_{\theta}(\hat{y}_i \mid x)}{P_{\theta_{\text{old}}}(\hat{y}_i \mid x)}
\end{equation}

The clipping factor $B$ is annealed with training step $s$:

\begin{equation}
B_s = \prod_{i=1}^{s} \left( \frac{s_{\text{total}} - i}{s_{\text{total}}} \right) \cdot \epsilon
\end{equation}

\paragraph{Dynamic Sampling.}
To avoid gradient collapse when all $G$ samples receive identical rewards, we skip such batches and continue sampling until variance is non-zero.

\vspace{1ex}
In summary, our hybrid reward design and GRPO optimization jointly enhance the model’s factual precision, causal reasoning clarity, and alignment with real-world moderation policies.

\section{Experiments}

\begin{figure*}[t]
    \centering
    \resizebox{\textwidth}{!}{%
        \begin{tabular}{@{}c@{\hspace{2pt}}c@{}} 
            \includegraphics[height=10cm, keepaspectratio, trim={0 0 5pt 0}, clip]{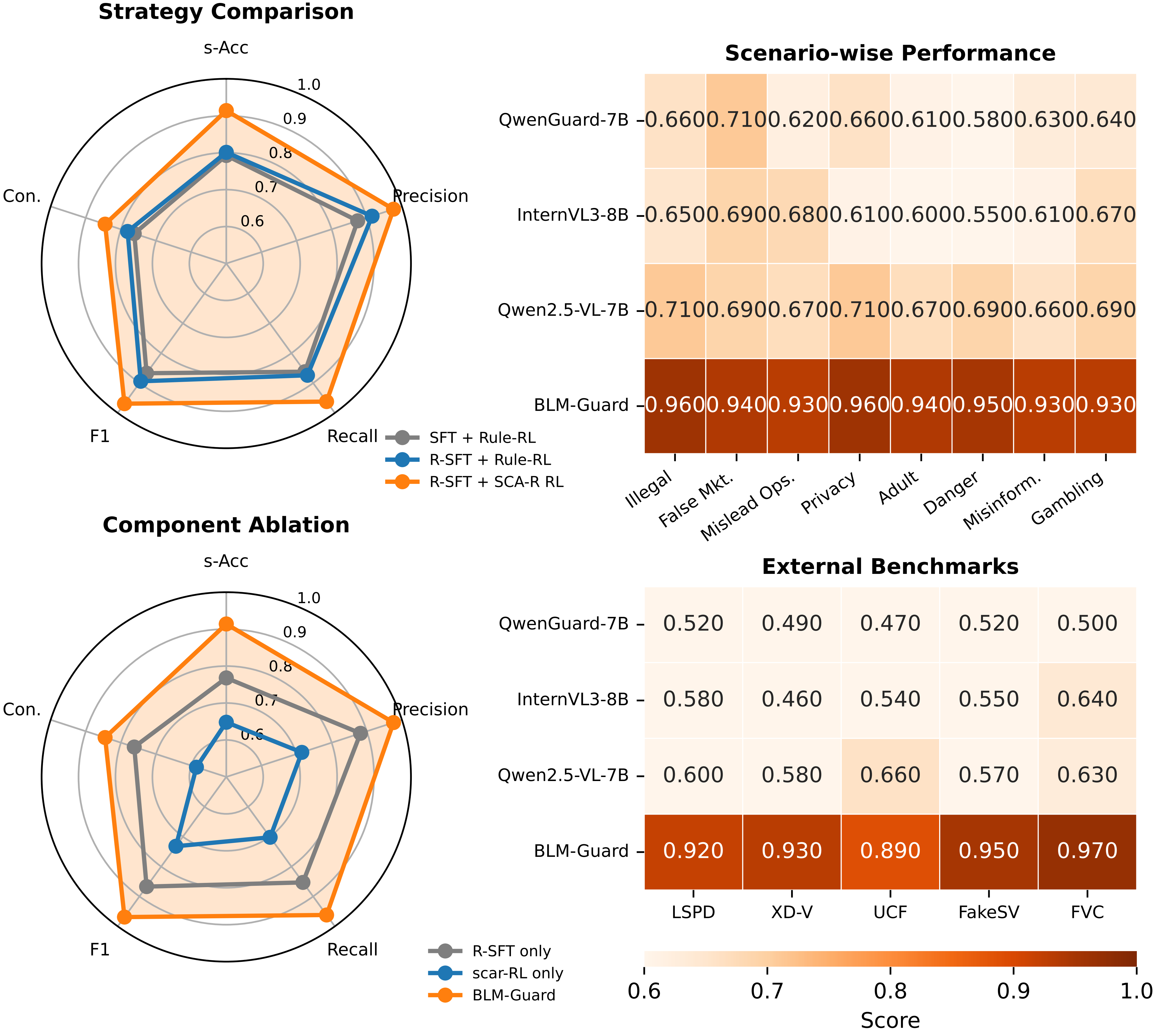} &
            \includegraphics[height=10cm, keepaspectratio, trim={5pt 0 0 0}, clip]{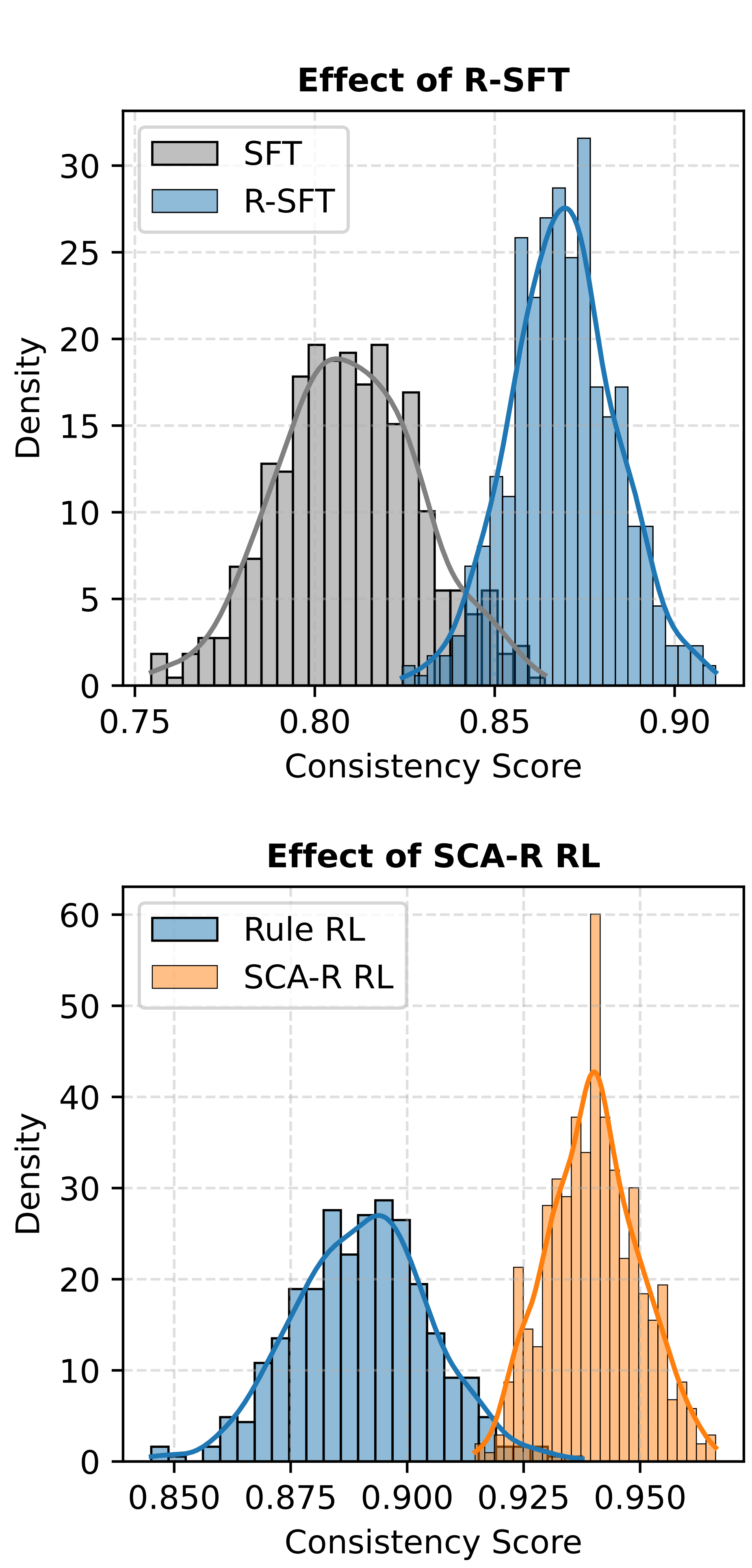}
        \end{tabular}%
    }
\caption{
    Comprehensive comparison of model strategies and components. \textbf{Left}: Radar plots highlight improvements in accuracy, precision, and consistency across key metrics. \textbf{Middle}: Heatmaps visualize performance across fine-grained risk scenarios and external benchmarks. \textbf{Right}: Histograms show the distributional gains in consistency induced by Rule-SFT and SCA-R reward learning.
    }
    \label{fig:experiment_results}
\end{figure*}

\paragraph{Baselines} 
We compare \textbf{BLM-Guard} against several state-of-the-art open-source multimodal models. These include guard-specific architectures \textbf{LlavaGuard-7B} and \textbf{QwenGuard-7B}~\cite{helff2025llavaguardopenvlmbasedframework}, and strong general-purpose foundation models: \textbf{LLaVA-Next-Video-7B}~\cite{zhang2024llavanextvideo}, \textbf{InternVL3-8B}~\cite{zhu2025internvl3exploringadvancedtraining}, and \textbf{Qwen2.5-VL-7B-Instruct}~\cite{bai2025qwen25vltechnicalreport} (our backbone). We also evaluate \textbf{Kimi-VL-A3B-Thinking}~\cite{kimiteam2025kimivltechnicalreport}, which utilizes structured Chain-of-Thought reasoning. For fair comparison, all models are adapted to video moderation via uniform frame sampling and ASR transcript concatenation.

\paragraph{Datasets.} We evaluate \textbf{BLM-Guard} on our \textbf{BLM-Guard-Bench}, a custom benchmark annotated with a three-level policy taxonomy: \textit{Severity}, \textit{Scenario}, and \textit{Violation Type} (Fig.~\ref{fig:benchmark}). This dataset includes structured reasoning traces for supervised and reward learning. To assess generalization, we further incorporate five public datasets: LSPD~\cite{phan2022lspd}, XD-Violence~\cite{wu2020not}, UCF-Crime~\cite{Sultani_2018_CVPR}, FakeSV~\cite{qi2023fakesv}, and FVC~\cite{papadopoulou2018fvc}, covering violence, anomaly, and misinformation detection. Evaluation of unseen policy domains follows the protocol in SafeWatch~\cite{chen2024safewatchefficientsafetypolicyfollowing}.


\paragraph{Metrics} We evaluate moderation along two dimensions:

\textbf{(1) Violation Identification Accuracy.} Based on structured \texttt{<answer>} outputs, we report \emph{Wide Accuracy} (partial match of violation scene or type) and \emph{Strict Accuracy} (full match of both). We also include macro/weighted precision, recall, and F1-score for harmful vs. harmless classification.

\textbf{(2) Consistency Score.} To assess reasoning quality in the \texttt{<think>} section, we use a GPT-4o-based guide model to score alignment with moderation principles. The score ranges from 0 to 1, with higher values indicating more consistent and interpretable reasoning.

\begin{table*}[htbp]
\centering
\footnotesize
\begin{adjustbox}{max width=\textwidth}
\begin{tabular}{l
                cccc
                cc
                ccccc
                c}
\toprule
& \multicolumn{4}{c}{\textbf{Severity-wise Classification}} 
& \multicolumn{2}{c}{\textbf{Accuracy}} 
& \multicolumn{5}{c}{\textbf{Binary Detection}} 
& \textbf{Con.} \\
\cmidrule(lr){2-5}\cmidrule(lr){6-7}\cmidrule(lr){8-12}\cmidrule(lr){13-13}
\textbf{Model} 
& \textbf{High} & \textbf{Medium} & \textbf{Low} & \textbf{None} 
& \textbf{w-Acc.} & \textbf{s-Acc.} 
& \textbf{NR (B)} & \textbf{R (B)} & \textbf{P} & \textbf{R} & \textbf{F1 (R)}
& \textbf{Con.} \\
\midrule
Qwen2.5-VL-7B            & 0.745 & 0.701 & 0.669 & 0.781 & 0.712 & 0.701 & 0.755 & 0.918 & 0.831 & 0.576 & 0.680 & 0.642 \\
Qwen2.5-VL-32B           & 0.732 & 0.685 & 0.652 & 0.768 & 0.703 & 0.682 & 0.875 & 0.823 & 0.769 & 0.835 & 0.801 & 0.667 \\
Kimi-VL-A3B-Thinking     & 0.728 & 0.684 & 0.651 & 0.624 & 0.529 & 0.511 & 0.887 & 0.561 & 0.588 & 0.898 & 0.711 & 0.701 \\
InternVL3-14B            & 0.581 & 0.535 & 0.502 & 0.622 & 0.521 & 0.499 & 0.931 & 0.519 & 0.485 & 0.950 & 0.642 & 0.593 \\
InternVL3-8B             & 0.695 & 0.645 & 0.607 & 0.701 & 0.389 & 0.631 & 0.921 & 0.299 & 0.461 & 0.961 & 0.623 & 0.611 \\
QwenGuard-7B             & 0.733 & 0.631 & 0.599 & 0.693 & 0.582 & 0.512 & 0.823 & 0.695 & 0.642 & 0.785 & 0.706 & 0.598 \\
LlavaGuard-7B            & 0.301 & 0.322 & 0.395 & 0.392 & 0.255 & 0.235 & 0.329 & 0.258 & 0.193 & 0.251 & 0.218 & 0.527 \\
LLaVA-Next-Video-7B      & 0.321 & 0.345 & 0.435 & 0.443 & 0.342 & 0.312 & 0.348 & 0.3101 & 0.1540 & 0.1780 & 0.165 & 0.533 \\

\textbf{BLM-Guard (ours)}& \textbf{0.978} & \textbf{0.940} & \textbf{0.902} & \textbf{0.986} & \textbf{0.962} & \textbf{0.914} & \textbf{0.781} & \textbf{0.950} & \textbf{0.976} & \textbf{0.962} & \textbf{0.969} & \textbf{0.845} \\
\bottomrule
\end{tabular}
\end{adjustbox}
\caption{Evaluation results of harmfulness detection for short-video advertisements. The left block reports severity-wise classification accuracy across \textit{High}, \textit{Medium}, \textit{Low}, and \textit{None} levels. The middle block presents overall accuracy, including lenient accuracy (\textbf{w-Acc.}) and strict accuracy (\textbf{s-Acc.}). The right block shows binary detection metrics: recall of non-risky (\textbf{NR (B)}) and risky (\textbf{R (B)}) classes, along with precision (\textbf{P}) and recall (\textbf{R}) for the risky class. “\textbf{Con.}” denotes reasoning–answer consistency scores assessed by GPT-4o (ranging from 0 to 1). Bold values indicate the best performance in each column; \textbf{BLM-Guard} achieves the strongest results across all metrics.}
\label{tab:ablation_main}
\end{table*}
\paragraph{Results on BLM-Guard Benchmark}
As Table~\ref{tab:ablation_main} shows, \textbf{BLM-Guard} outperforms all baselines in severity classification, overall accuracy, binary risk detection, and reasoning consistency, producing more stable and principle-aligned outputs. This demonstrates the effectiveness of combining rule-guided supervision, structured reasoning, and reinforcement learning.

\paragraph{Generalization on Existing Datasets.}
\textbf{BLM-Guard} generalizes well to public benchmarks (LSPD, XD-Violence, UCF-Crime, FakeSV, FVC), particularly excelling in misinformation scenarios (FakeSV/FVC) where prior models struggle. While these datasets do not directly evaluate factual consistency, \textbf{BLM-Guard} effectively detects deceptive content and cross-modal inconsistencies, consistent with its internal benchmark performance (see Figure~\ref{fig:experiment_results}).

\section{Ablation Study}
\label{sec:ablation}

We evaluate \textsc{BLM-Guard} (Qwen2.5-VL-7B) across seven configurations to verify the impact of our training stages: 
(1) \textbf{Ans-SFT} and (2) \textbf{Think-SFT} (single-component tuning); 
(3) \textbf{Full-SFT (13k)} vs. (4) \textbf{Part-SFT (5k)} (data scaling); 
(5) \textbf{Rule-SFT (5k)} (rule-anchored joint tuning); 
(6) \textbf{+Rule-RL} (rule-based reward learning); 
and (7) \textbf{+SCA-R (Full)}, our final setup with SCA-R guided RL. 
The results, summarized in Table~\ref{tab:ablation_test}, highlight the synergy between rule-aligned SFT and SCA-R refinement.

\begin{table}[H]
\centering
\setlength{\tabcolsep}{3.5pt} 
\renewcommand{\arraystretch}{1.0}
\small

\begin{tabular}{lccccc}
\toprule
\textbf{Setting} & \textbf{s-Acc.} & \textbf{P} & \textbf{R} & \textbf{F1} & \textbf{Con.} \\
\midrule
Ans-SFT                  & 0.648  & 0.765 & 0.702 & 0.732 & 0.412 \\
Think-SFT                & 0.612  & 0.720 & 0.679 & 0.699 & 0.585 \\
Full-SFT (13k)           & 0.7925 & 0.874 & 0.862 & 0.867 & 0.762 \\
Part-SFT (5k)            & 0.768  & 0.780 & 0.840 & 0.809 & 0.762 \\
Rule-SFT (5k)            & 0.783  & 0.882 & 0.853 & 0.867 & 0.776 \\
\quad + Rule-RL          & 0.801  & 0.915 & 0.874 & 0.894 & 0.781 \\
\quad + SCA-R (Full)     & \textbf{0.914}  & \textbf{0.976} & \textbf{0.962} & \textbf{0.969} & \textbf{0.845} \\

\bottomrule
\end{tabular}
\caption{
Ablation on Qwen2.5-VL-7B. \textbf{Ans-SFT}/\textbf{Think-SFT} are single-supervision baselines. \textbf{Rule-SFT} uses rule-aligned tuning. \textbf{+Rule-RL}/\textbf{+SCA-R} introduce reward learning; SCA-R adds adaptive alignment. \textbf{Con.} denotes reasoning consistency.
\label{tab:ablation_test}
}

\label{tab:ablation_qwen7b}
\end{table}
\vspace{0.5mm}
\noindent \textbf{Findings:}  
(1) Rule-SFT outperforms single-stage SFT by improving both accuracy and interpretability.  
(2) RL enhances robustness; combining Rule-SFT with rule-based RL yields strong gains.  
(3) The full version (+SCA-R) achieves the best overall performance in reasoning quality and policy alignment.

\section{Related Work}
\label{sec:related}

\paragraph{VLM Guardrails.} 
Early guardrails like \textsc{LlamaGuard}~\cite{inan2023llamaguardllmbasedinputoutput} and \textsc{LlavaGuard}~\cite{helff2025llavaguardopenvlmbasedframework} focused on text/image safety. While recent works extend to video~\cite{chen2024safewatchefficientsafetypolicyfollowing} and legal alignment~\cite{lu2025vlmpolicycommonlawcontent}, they often fail to capture nuanced ad violations like misleading visuals or exaggerated claims.

\paragraph{CoT and RL Alignment.} 
CoT interpretability~\cite{tanneru2024hardnessfaithfulchainofthoughtreasoning, xu-etal-2024-faithful} and multi-step reasoning (e.g., Tree-of-Thoughts~\cite{yao2023treethoughtsdeliberateproblem}, GuardAgent~\cite{xiang2025guardagentsafeguardllmagents}) are foundational to our approach. In multimodal safety, GuardReasoner~\cite{liu2025guardreasonerreasoningbasedllmsafeguards} and MM-RLHF~\cite{zhang2025mmrlhfstepforwardmultimodal} explore preference-based RL. We extend these by incorporating policy-specific, dynamic principles to boost alignment.

\paragraph{Rule-Guided Data \& ICoT.} 
Instruction tuning (e.g., Self-Instruct~\cite{wang2023selfinstructaligninglanguagemodels}) and policy-grounded generation (e.g., MSR-Align~\cite{xia2025msralignpolicygroundedmultimodalalignment}) minimize annotation costs. Unlike these, we introduce Interleaved CoT (ICoT) and adaptive keyframe sampling specifically designed for structured reasoning in multimodal ad contexts.

\section{Conclusion}

We propose \textbf{BLM-Guard}, an explainable multimodal framework for short-video ad moderation. By integrating \emph{Chain-of-Thought} (CoT) reasoning with rule-aligned Reinforcement Learning (RL), BLM-Guard ensures high accuracy and policy-grounded transparency. Leveraging a risk-tiered benchmark, the framework utilizes a multi-task objective to detect nuanced violations, such as intra-modal exaggerations and cross-modal mismatches. Experiments show that BLM-Guard significantly outperforms SOTA baselines and exhibits robust generalization across unseen policies and diverse datasets, proving its utility in high-stakes moderation.

\bibliography{references/safe_guard,references/post_train,references/interpret}

\end{document}